# TflosYOLO+TFSC: An Accurate and Robust Model for Estimating Flower Count and Flowering Period

**Authors**

Qianxi Mi1,2, Pengcheng Yuan1,2, Chunlei Ma1,2, Jiedan Chen1,2*, Mingzhe Yao1,2*

**Affiliations**

1 Key Laboratory of Biology, Genetics and Breeding of Special Economic Animals and Plants, Ministry of Agriculture and Rural Affairs, Tea Research Institute of the Chinese Academy of Agricultural Sciences, Hangzhou 310008, China.

2 National Key Laboratory for Tea Plant Germplasm Innovation and Resource Utilization, Tea Research Institute of the Chinese Academy of Agricultural Sciences, Hangzhou 310008, China.

*Address correspondence to: Jiedan Chen (chenjd@tricaas.com); and Mingzhe Yao (yaomz@tricaas.com).

**Abstract**

Tea flowers play a crucial role in taxonomic research and hybrid breeding for the tea plant. As traditional methods of observing tea flower traits are labor-intensive and inaccurate, we propose TflosYOLO and TFSC model for tea flowering quantifying, which enable estimation of flower count and flowering period. In this study, a highly representative and diverse dataset was constructed by collecting flower images from 29 tea accessions in 2 years. Based on this dataset, the TflosYOLO model was built on the YOLOv5 architecture and enhanced with the Squeeze-and-Excitation (SE) network, which is the first model to offer a viable solution for detecting and counting tea flowers. The TflosYOLO model achieved an mAP50 of 0.874, outperforming YOLOv5, YOLOv7 and YOLOv8. Furthermore, TflosYOLO model was tested on 34 datasets encompassing 26 tea accessions, five flowering stages, various lighting conditions, and pruned/unpruned plants, demonstrating high generalization and robustness. The correlation coefficient ($R^2$) between the predicted and actual flower counts was 0.974. Additionally, the TFSC (Tea Flowering Stage Classification) model – a 7-layer neural network was designed for automatic classification of the flowering period. TFSC model was evaluated on 2 years and achieved an accuracy of 0.738 and 0.899 respectively. Using the TflosYOLO+TFSC model, we monitored the tea flowering dynamics and tracked the changes in flowering stages across various tea accessions. The framework provides crucial support for tea plant breeding programs and phenotypic analysis of germplasm resources.

1. **INTRODUCTION**

Tea is one of the three major beverages in the world, and the tea plant is an important economic crop in multiple countries. With a cultivation history spanning thousands of years, China is home to a rich diversity of native tea accessions. In recent years, numerous distinct tea cultivars have been developed across various tea-growing regions, supporting the growth of the tea industry and promoting improvements in both quality and efficiency. As a perennial leaf crop, the economic value of tea plant primarily derives from its young shoots, and most research has focused on the growth and development of these shoots. However, as reproductive organs, tea flowers are crucial for conducting genetic and taxonomic studies. The flowering period is crucial for selecting parent plants for hybrid breeding, as it must be relatively synchronized for successful cross-breeding. Tea flowering consumes the plant nutrients so that flower thinning can regulate carbon-nitrogen metabolism, promoting vegetative growth while suppressing reproductive growth, further enhancing the yield of young shoots and increasing the amino acid content, which positively impacts tea quality[1,2]. Therefore, measuring the floral phenotypes of tea accessions is of great importance.

China has abundant phenotypic resources of tea plants, and significant differences exist between accessions in terms of flower quantity and flowering stage (including the onset and cessation of blooming, and the duration of the flowering stage). Breeding programs require investigations of flower quantity and flowering stages. However, Traditional methods for observing tea flower traits, such as manual measurements, are labor-intensive and prone to inaccuracies. Additionally, given the diversity of tea accessions, with differences in flower size, color, quantity, and flowering period, previous studies have only selected a small number of accessions[1], making it difficult to accurately describe the regional characteristics of the species. Therefore, there is a clear need to develop efficient, precise, and highly generalized phenotyping technologies for tea flowers.

In recent years, advancements in machine learning, deep learning, computer vision technologies, and drones have significantly impacted agricultural applications, such as yield prediction, crop growth monitoring, automated harvesting, and quality detection. Traditional machine learning methods (ML), including support vector machines (SVM), random forests, partial least squares regression (PLSR), K-means clustering, and artificial neural networks (ANN), take a data-driven approach to model the relationships between input data and labels, such as crop yield[3]. These machine learning systems are capable of processing large datasets and handling non-linear tasks efficiently[4]. For example, A machine learning algorithm incorporating K-means clustering was developed for grapevine inflorescence detection, classification, and flower number estimation, which demonstrates high accuracy[5]. In another study, six different machine learning algorithms, including ridge regression, SVM, random forest, Gaussian process, K-means, and Cubist was utilized by Song et al.[6] to establish yield prediction models, based on drone-collected visible and multispectral images of wheat canopies during the grain filling stage. As for machine learning in tea research, Tu et al.[7] utilized UAV-acquired hyperspectral data to build a classification model for tea accessions and estimate the content of key chemicals related to tea flavor. Their research indicated that SVM and ANN models were most effective for tea plant classification. Chen et al.[8] compared the performance of multilayer perceptron (MLP), SVM, random forests (RF), and PLSR using hyperspectral data from tea plants, developing Tea-DTC model for evaluating drought resistance traits in 10 tea plant germplasm resources.

However, traditional machine learning methods are heavily reliant on manually selected features under controlled conditions, and their robustness tends to be limited, particularly in complex field environments. These methods often struggle to handle the challenges posed by the dynamic and variable real-world agricultural environments[9,10]. Deep learning (DL) methods, on the other hand, excel in discovering patterns and hidden information from large datasets using neural networks[11]. Unlike traditional machine learning, DL approaches are better suited for complex scenarios and require large amounts of data for training. Recent deep learning algorithms, such as Faster R-CNN, ResNet, and YOLO-based models, have demonstrated superior performance in crop yield estimation[12,13], growth monitoring[14,15], and object detection for fruits and other crop targets[16–18]. Additionally, the integration of machine learning, deep learning, and plant phenotyping platforms, along with UAV technology, has resulted in the development of many new and efficient techniques. For instance, RGB and multispectral images was utilized to identify the tasseling stage of maize[21]. Drone time-series images and a Res2Net50 model was used to identify five growth stages of rice germplasm using, achieving good prediction results for the heading and flowering stages by combining RGB and multispectral images and developing a PLSR model[22]. Similarly, drone time-series images and deep learning models were applied for dynamic monitoring of maize ear area[23]. These advances have significantly contributed to the rapid and efficient extraction of plant information, facilitating accurate plant phenotyping.

YOLOv5, developed by Glenn Jocher et al.[19], is an improved version of YOLOv3. It is characterized by a relatively small model size and fast processing speed, making it suitable for mobile deployment. In recent years, YOLO-based algorithms, particularly YOLOv5 have been widely applied to object detection in agriculture, demonstrating superior performance on agricultural datasets[20].

Several automatic detection models for various flowers such as apple flowers, pear flowers, grapevine flowers, strawberry flowers and litchi flower, have been developed[5,10,24–28], as well as tea shoot detection models[29–33]. For instance, Wang et al.[34] used color thresholding followed by SVM classification to estimate mango inflorescence area, employing Faster R-CNN for panicle detection. Lin et al.[24] proposed a framework for counting flowers in Litchi panicles and quantifying male Litchi flowers, employing YOLACT++ for panicle segmentation and a novel algorithm based on density map regression, for accurate flower counting. The YOLOX was utilized by Xia et al.[10] for tree-level apple inflorescence detection, achieving the highest AP50 of 0.834 and AR50 of 0.933.

To date, however, no models have been specifically developed to detect tea plant flowers or observe tea flower phenotypes. To fill this gap, we propose a method for tea flowering quantifying, comprising the TflosYOLO model and TFSC (Tea flowering stages classification) model. TflosYOLO model based on YOLOv5, is the first to offer a viable solution for detecting and counting tea flowers, with potential applications in tea flower thinning practices. TFSC model is a novel model for Tea flowering stages classification, which can detect the flowering period after TflosYOLO model. The framework is designed to enable dynamic monitoring of flower quantity and flowering periods across various tea accessions, providing crucial support for tea plant breeding programs and phenotypic analysis of germplasm resources.

## 2. MATERIALS AND METHODS

### 2.1 Experimental Design

Estimating Flower Count and Flowering Period is achieved through time-series images of tea flowers, TflosYOLO and TFSC model. The framework as shown in Fig. 1. The process is outlined as fronts: Mobile phone images of tea plant flowers are captured to establish a tea flower dataset, which is then used to train the TflosYOLO model. TflosYOLO model provide the detection results for tea flower buds (bud), blooming flowers (B-flower), and withered flowers (W-flower), which are then used to output flower counts. The tea flowering stage classification model is used to determine the flowering stage (IFS, EFS, MFS, LFS, TFS).

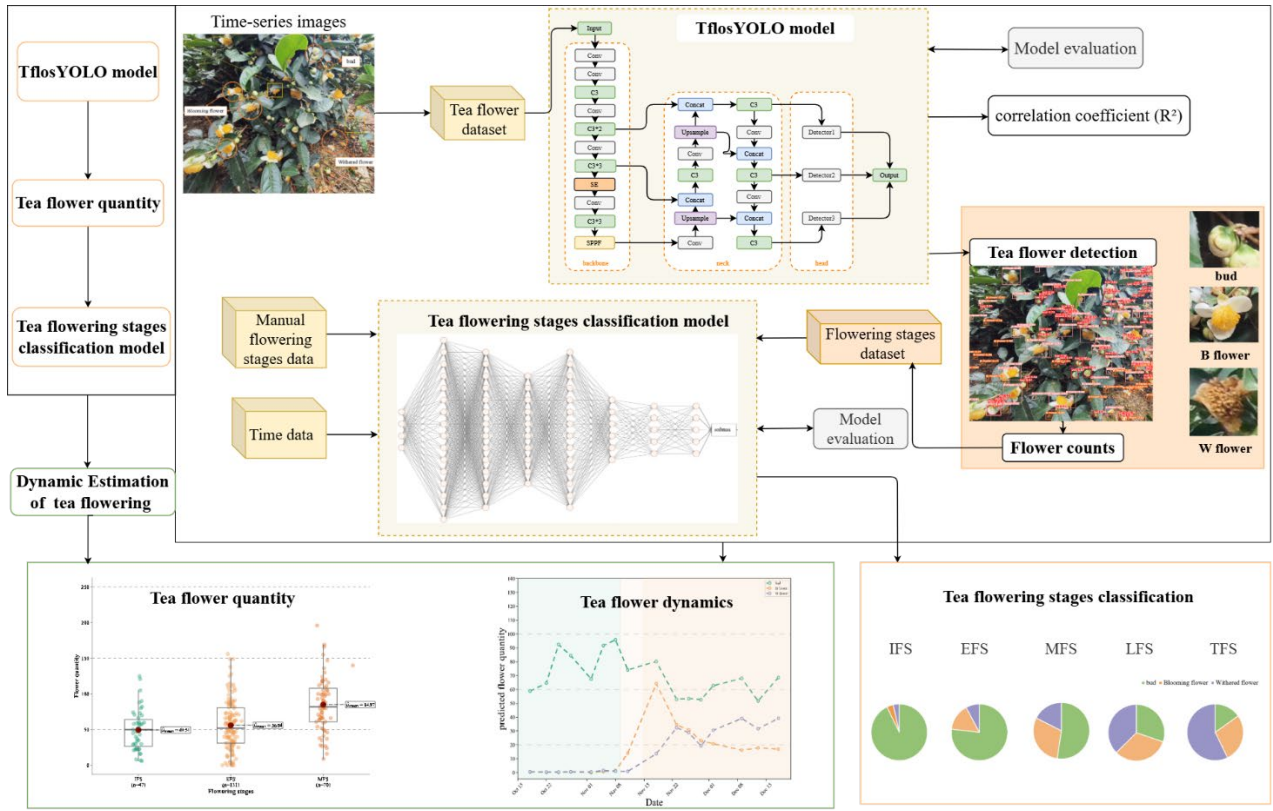

Fig. 1. Overall framework for dynamic estimation and analysis of tea flowering

**2.2 Study site and materials**

The experimental data used in this study were obtained in November - December 2023 and October - December 2024 at the National Tea Germplasm Research Garden (Hangzhou). Hangzhou is located in the southeastern region of China (29°–30°N and 118°–120°E), within a subtropical monsoon climate zone. Our research involved 29 tea accessions originating from different regions across the country, information on these accessions is provided in Table S1, S2.

**2.3 Data acquisition**

Tea plants in tea gardens are typically planted in rows with dense spacing between individual plants. Their flowers generally bloom predominantly on the sides of the plants. Considering this, we utilized mobile phone for image capture, as mobile phone photography offers flexibility, making it feasible for large-scale, cost-effective, and precise phenotypic monitoring. The mobile phone capture images in RGB color format as JPG files. The image resolution is 3280×2464 pixels, with a 72 dpi setting. The actual area corresponding to the regions captured in each image was calculated using Fiji[35] and is approximately 3690.33 cm$^2$ (69.26cm × 53.28cm per image), detailed method is shown in Figures S1. In order to enhance the generalization ability of the model, images in the complexity environments were collected in 2023-2024, including different lighting (e.g., backlight and frontlight), 29 tea accessions, various flowering densities, and both pruned and unpruned tea plants. In total, over 14,151 images were token.

To evaluate the reliability of our approach as a substitute for traditional manual measurement and to explore the relationship between manual investigations and this framework, we conducted manual assessments of flower quantity and flowering stages after every image collection. The method of manual assessments is illustrated in Fig. S2.

## 2.4 Images annotation and dataset analysis

### 2.4.1 Images annotation

The original images captured by mobile phones had a resolution of 3280×2464 pixels. The input image size for the YOLO model was determined based on the specific model configuration and task requirements. In this study, the input size for model training, validation, and testing was set to 640×640 pixels. To ensure compatibility with this input size and reduce computational cost, the original images were cropped into four sub-images, each with a size of 1640×1232 pixels. Image annotation was performed using LabelImg[36] in YOLO format. The labeled images were divided into three datasets for training, validation, and testing, following a 6:2:2 ratio. Three categories were defined for annotation: buds, blooming flowers (B flower), and withered flowers (W flower) (Fig. S3). In total, 28,668 instances were labeled across 2,361 images in the tea flower dataset.

Additionally, various additional test datasets were constructed after annotation to assess the model performance.

*2.4.2 Dataset for TflosYOLO model*

Three datasets were constructed for the training, validation, and testing of the tea flower detection model. The final annotated dataset included 2,361 images with a total of 28,668 instances: 57% were buds, 25% were B flower, and 18% were W flower. Buds accounted for the majority of the instances, while withered flowers represented only 18%, indicating a class imbalance in tea flower dataset (Table. 1, Fig. S4). To ensure reliability, generalization capability, and robustness of tea flower detection model, the tea flower dataset includes images from 26 tea accessions originating from six provinces of China (Table S1).

Table. 1. The amount of different classes in tea flower dataset.

| class | Training set (1432 images) | Validation set (469 images) | Test set (460 images) | All instances |
|---|---|---|---|---|
| bud | 9447 | 3300 | 3645 | 16392 |
| Blooming flower | 4303 | 1410 | 1538 | 7251 |
| Withered flower | 2905 | 996 | 1124 | 5025 |
| All class | 16655 | 5706 | 6307 | 28668 |

Moreover, 34 additional test datasets were constructed to evaluate the model on various tea accessions, flowering stages, lighting conditions (backlight and front light), and unpruned tea plant images. Except unpruned test set, all test dataset constructed by pruned tea plants images. The representative images and the amount of images for 34 additional test datasets have been provided in Fig. S5, Table S3.

**2.5 TflosYOLO model for tea flower detection**

*2.5.1 TflosYOLO model and YOLOv5*

Although YOLOv7[37] and other YOLO model has also shown excellent performance on agricultural datasets, considering the trade-off between model accuracy and computational cost, we adopt YOLOv5m as the baseline model for further improvement, aiming to achieve accurate and efficient tea flower detection across various environments and accessions while minimizing computational costs. TflosYOLO can be regarded as an new version of YOLOv5, which is more suitable for flower detection and has additional function for direct flower counting.

The YOLOv5 network consists of three main components: (a) Backbone: CSPDarknet, (b) Neck: PANet, (c) Head: YOLO Layer. Initially, data is passed through the CSPDarknet for feature extraction. Next, it is processed through PANet to achieve feature fusion. Finally, the YOLO layer performs object detection and classification, outputting the final results in terms of detected objects and their corresponding classes.

In the detection process of YOLO-based algorithms, the input image is processed to generate feature map, which is divided into an S×S grid. For each grid cell, anchor boxes are scored and boxes with low scores are discarded. Non-Maximum Suppression (NMS) is then applied to eliminate redundant boxes. Only the remaining boxes, along with their confidence scores, are retained and displayed. The confidence score is calculated as:

$$\text{confidence score} = \Pr(\text{object}) * \text{IoU}(\text{pred}, \text{truth}) * \Pr(\text{class}) \qquad (1)$$

Where:
- $\Pr(\text{object})$ represents the probability that an object exists,
- IoU represents the Intersection over Union between the predicted and ground truth boxes,
- $\Pr(\text{class})$ represents the probability that the predicted box belongs to each class.

IoU is Area of Intersection. The IoU is calculated as:

$$\text{IOU} = \frac{\text{area}(B_P \cap B_{gt})}{\text{area}(B_P \cup B_{gt})} \qquad (2)$$

$B_P$ is predicted bounding box, $B_{gt}$ is ground truth box.

The YOLOv5 loss function consists of three components: classification loss, objectness loss, and box loss. To compute the total loss, these three components are combined as a weighted sum, which is expressed as follows:

$$\text{Loss} = w_{box}l_{box} + w_{obj}l_{obj} + w_{cls}l_{cls} \tag{3}$$

- $l_{box}$ is the box regression loss, which measures the difference between the predicted and ground truth box locations,
- $l_{box}$ is the object confidence loss, which evaluates the accuracy of the model's object detection,
- $l_{cls}$ is the classification loss, which measures the model ability to classify the detected objects accurately.

2.5.2 Challenges in tea flower detection

There are multiple challenges in tea flower detection as shown in Fig. 2. The field environment of tea garden is complex, with varying light conditions, backgrounds, and other factors contributing to significant background noise. In addition to this, tea flowers are small and tend to grow on the side of the tea plant densely, with buds and flowers often obscuring each other, prone to being obstructed or fragmented by branches and leaves, and they are also easily influenced by background flowers. These factors make tea flower detection more challenging compared to the detection of fruits like apples[10]. Furthermore, intermediate forms exist between buds, blooming flowers, and withered flowers, which are difficult to differentiate and can lead to a decrease in detection accuracy. Additionally, light interference, such as light spots, can cause buds to be misidentified. The imbalance among different flower categories is also one of the challenges as the total number of tea buds and blooming flowers is significantly greater than the number of withered flowers. To address these challenges, this study proposes the TflosYOLO model, which aims to improve the accuracy of tea flower detection under various environmental conditions.

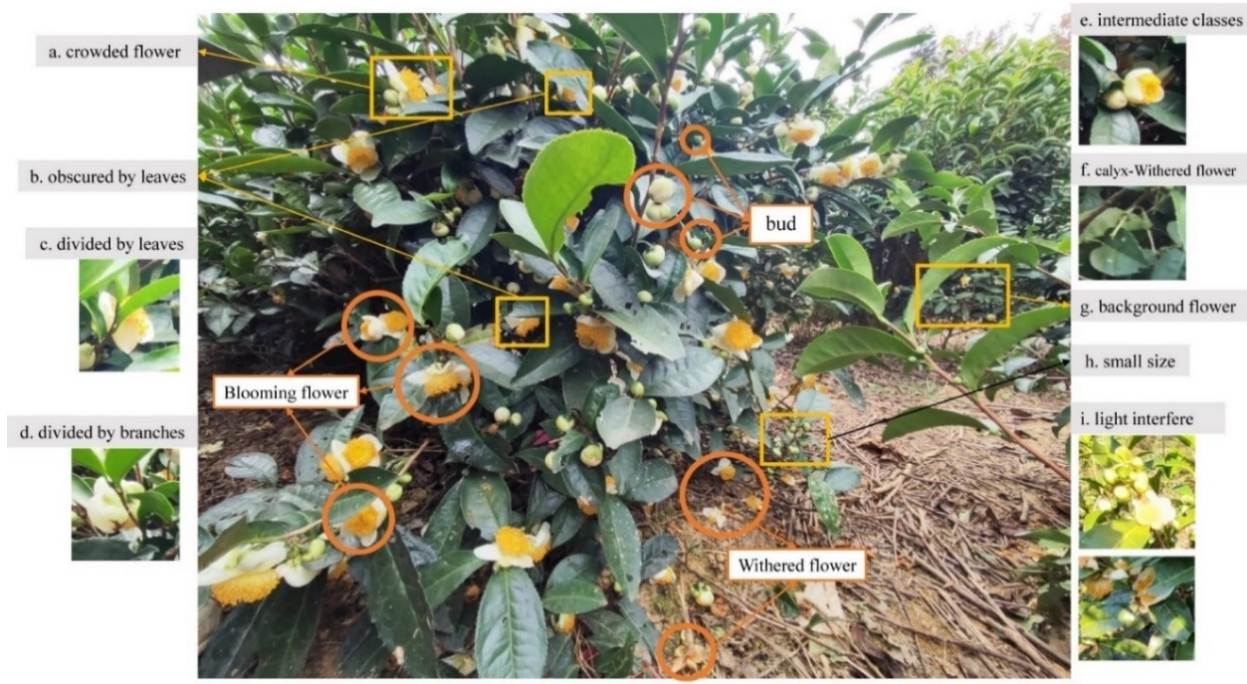

Fig. 2. Examples of the inflorescences on tea plant and difficult issues in tea flower detection. (a) crowded flower obscured by each other. (b) tea flower obscured by leaves. (c) tea flower divided by leaves. (d) tea flower divided by branches. (e) intermediate classes. (f) calyx belong to Withered flower, which can be easily detected as bud. (g) background flower that do not belong to the detected tree. (h) small size detection target. (i) light interfere.

2.5.3 Architecture of TflosYOLO model

In this study, the YOLOv5m model is used as the baseline. Modifications to the model's depth and width are made, and the SE (Squeeze-and-Excitation Networks) attention module is integrated into the backbone of YOLOv5. And we add the additional function to output the flower counts directly as CSV. After improvement, TflosYOLO model is more suitable for flower detection and flower counting.

The SE module is added to the seventh layer of the YOLOv5 backbone, which enhances the model ability to handle complex backgrounds and lighting variations. The TflosYOLO model achieves high tea flower detection accuracy with relatively low computational cost, offering excellent generalization and robustness. The architecture of the TflosYOLO model (Fig. 3A) includes the backbone (CSPDarknet-53), the feature fusion neck, and the final detection layers.

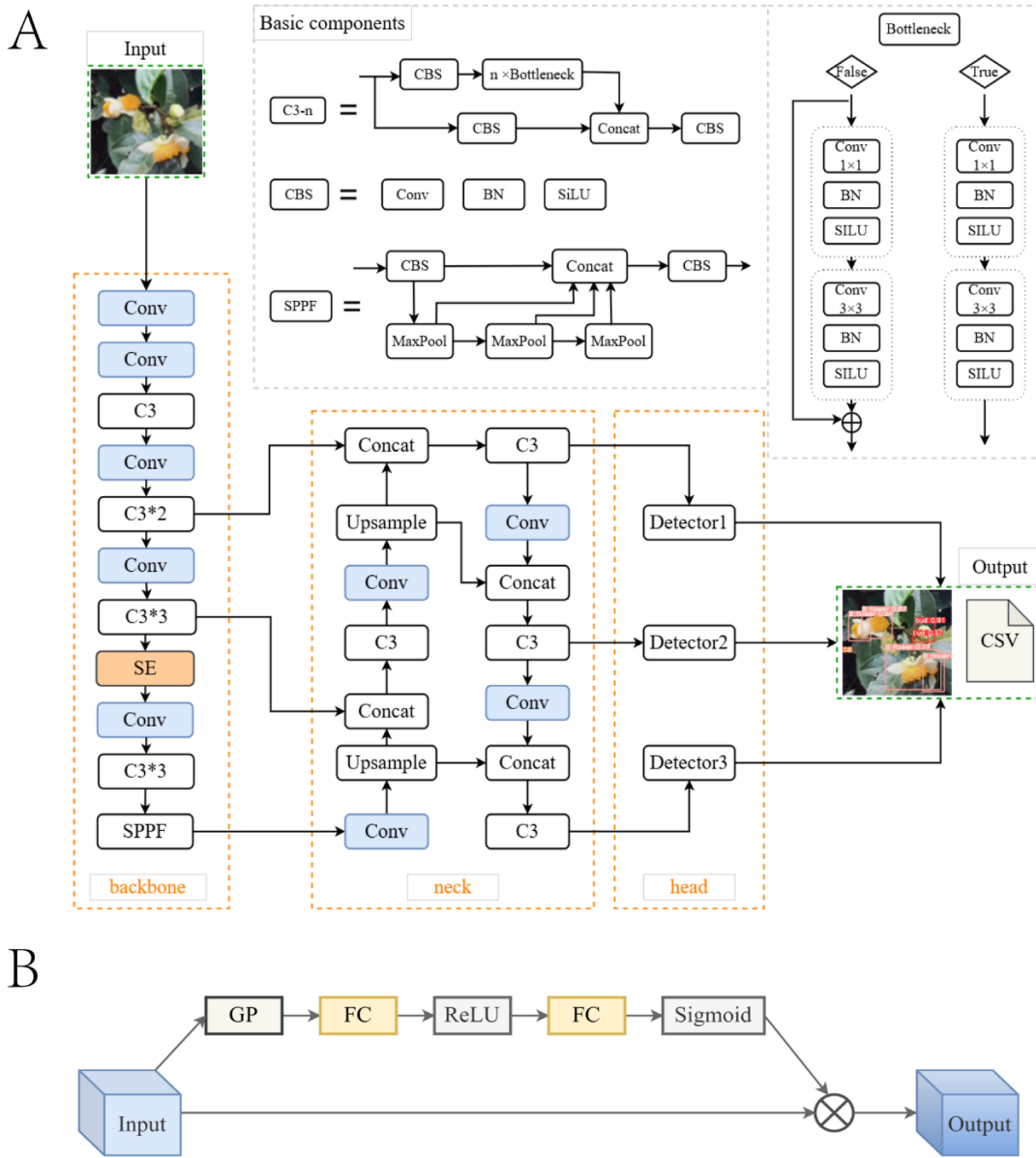

Fig. 3. The model structure of TflosYOLO model. (A) The architecture of the TflosYOLO model. (B) The structure of SE (Squeeze-and-Excitation Networks). GP = global pooling, FC = fully-connected layer.

The images are input into the TflosYOLO model, with the input size scaled to 640×640(input size for detection varies depending on image size). The images pass through the main feature extraction network of the TflosYOLO model, generating various feature maps. These feature maps undergo further subsampling and feature fusion in the neck section, integrating shallow and deep features. The C3 modules at layers 18, 21, and 24 output feature maps of sizes 80×80, 40×40, and 20×20, respectively, for detecting small, medium, and large targets. The model divides the image into grids and generates anchor boxes of varying sizes and densities. Anchor boxes with high scores

(including both object and category scores) are retained, and non-maximum suppression (NMS) is applied to eliminate redundant anchor boxes. The remaining anchor boxes are displayed on the image along with their predicted class confidences, providing the detection results for bud, B-flower, and W-flower. Moreover, TflosYOLO model outputs flower quantities of 3 type of tea flower as CSV format for further analysis.

2.5.4 Key Improvements in the TflosYOLO Model

This model introduces Data Augmentation, YOLOv5 Model Scaling, integration of the SE Module and direct counting outputs. TflosYOLO can be regarded as an new version of YOLOv5 for better flower prediction and flower counting.

The training images are augmented with mosaic, flipping, translation, and color enhancement techniques to address the problem of insufficient training data, particularly for withered flower samples, which could lead to model underfitting. The YOLOv5 model includes several variants (YOLOv5s, YOLOv5m, YOLOv5l, YOLOv5x), all with identical model structures but differing depth_multiple and width_multiple parameters. The depth_multiple controls the number of modules, while width_multiple adjusts the number of convolution kernels to control the number of channels. In this study, we adjust both parameters, resulting in a model size between YOLOv5s and YOLOv5m. Additionally, the SE(Squeeze-and-Excitation Networks) module- a channel attention mechanism[38,39], is added to the seventh layer of the YOLOv5 model. The structure of SE is shown in Fig. 3B. The SE module consists of two key steps: Squeeze and Excitation. It dynamically adjusts the weights of different channel by learning the relationships between channels, in order to make the network focus on more important features while suppressing unimportant channels.

2.5.5 Training Details

The model was trained for 300 epochs with a batch size of 8 and a learning rate of 0.01, using the SGD optimizer. The input image is resized to 640×640 pixels. The experimental setup and Environmental settings are detailed in Table 2.

Table 2. experimental setup and Environmental settings.

| operating system | ubuntu18.04 |
|---|---|
| GPU | RTX 3080(10GB) *1 |
| CPU | Intel(R) Xeon(R) Platinum 8255C |
| version | pytorch-cuda=11.8, Cuda 11.3, Python 3.8 |

*2.5.6 model evaluation*

In order to assess the model for tea flower detection, eight key performance indicators (KPIs) are adopted in this study. Precision and Recall are commonly used evaluation metrics in deep learning Algorithm Evaluation, all of which are based on the confusion matrix[40]. The confusion matrix is presented in Table S4.

Precision is the proportion of True Positive (TP) in all detection- predicted positive samples (TP + FP). The formula is as fronts:

$$\text{Precision} = \frac{TP}{TP+FP} = \frac{TP}{\text{all detections}} \quad (4)$$

Recall is the proportion of True Positive (TP) in all actual positive samples (TP + FN). The formula is as fronts:

$$\text{Recall} = \frac{TP}{TP+FN} = \frac{TP}{\text{all actual positive}} \quad (5)$$

F1-score combines Precision and Recall to measure the performance of a model. The formula is as fronts:

$$F1 = 2 \times \frac{P \times R}{P+R} \quad (6)$$

Where R is Recall and P is Precision, C denotes class.

In object detection algorithms, Intersection over Union (IoU) is a commonly used metric to evaluate the accuracy of predicted bounding boxes against ground truth boxes. The formula is as fronts:

$$IOU = \frac{area(B_P \cap B_{gt})}{area(B_P \cup B_{gt})} \qquad (7)$$

Where $B_p$ is predicted bounding box, $B_{gt}$ is ground truth box.

Average Precision (AP) is a key metric used to assess the performance of detection models over one class, reflecting the trade-off between precision and recall. Specifically, mAP (mean Average Precision) averages the AP across different classes, mAP0.5 refers to the mAP calculated at an IoU threshold of 0.5; mAP0.5-0.95 represents the mean Average Precision calculated across a range of IoU (Intersection over Union) thresholds from 0.5 to 0.95. Formulas are provided below:

$$AP = \int_0^1 P(R) dR \qquad (8)$$

$$mAP = \frac{\sum_{n=0}^{c} AP(C)}{C} \qquad (9)$$

Where R is Recall and P is Precision, C denotes class.

Additionally, detection speed is used to evaluate detection time cost, while total parameters, FLOPs, and model size are crucial for evaluating model complexity and computational cost.

In this study, we use $R^2$ coefficient to assess the strength of the correlation between the manually observed, annotated, and predicted tea flower numbers, further validating the reliability of tea flower detection model. Formula for $R^2$ calculating are provided below:

$$R^2 = 1 - \frac{\sum_{i=1}^{n}(y_i - \hat{y}_i)^2}{\sum_{i=1}^{n}(y_i - \bar{y})^2} \qquad (10)$$

where n is the number of samples, $y_i$ is the manually observed or annotation flower quantity, and $\hat{y}_i$ is the predicted tea flower quantity from deep learning model, and $\bar{y}$ is the average of $y_i$.

## 2.6 Tea flowering stages classification model

This study constructs a tea flowering stages classification model -TFSC model, we used 7-layer neural network and time-series images to enable precise dynamic estimation of the flowering period.

### 2.6.1 Flowering stage dataset construction

The tea plant flowering stage is categorized into five stages: Initial Flowering Stage (IFS), Early Peak Flowering Stage (EFS), Mid Peak Flowering Stage (MFS), Late Peak Flowering Stage (LFS), and Terminal Flowering Stage (TFS). To construct the training and validation datasets, we utilized uncropped raw images of tea flowers collected in 2023. As the flowering periods of tea plants are influenced by climatic factors and can vary significantly between years, we incorporated tea flower images collected in 2024 to establish the test dataset. This test dataset, comprising 387 samples, aims to further validate the accuracy and generalizability of the flowering stage detection model.

Using the TflosYOLO model, we estimated the corresponding flower counts (including the number of flower buds, B-flower, and W-flower) for each image. Additionally, time data was incorporated. Manually recorded flowering stages were used as labels. Each image's flower quantity, manually observed flowering stage and time data, constituted a flowering stage sample, collectively forming the original flowering stage dataset.

Subsequently, we preprocessed the original flowering stage dataset by first filtering out low-quality data. This involved removing images of varieties with insufficient flower counts, as they could not provide reliable flowering stage assessments. For the remaining samples from the same time and accession, we calculated the average value from every three samples to create a new sample. This approach mitigates the influence of extreme cases and reflects the overall flowering characteristics of the accession. Each sample was then manually labeled with tags that included IFS, EFS, MFS, LFS, and TFS. The 2023 flowering period data was divided into training and validation sets in an 8:2 ratio, while the 2024 images served as the test set.

2.6.2 TFSC model design and training

The Flowering stage classification model is built using 7-layer neural network and the flowering period dataset. ANN, also known as Multilayer Perceptron (MLP), consist of fully connected layers. Each layer contains multiple artificial neural units (neurons). The model is implemented using the

PyTorch, with ReLU activation functions, softmax for classification, cross entropy loss and the Adam optimizer. Training parameters are shown in Table 3.

Table 3. Key training parameters.

| Training samples | 3667 |
|---|---|
| validation samples | 671 |
| Test samples | 387 |
| Batch size | 16 |
| Learning rate | 0.001 |
| Epochs | 80 |
| Software version | pytorch-cuda=11.8、Cuda 11.3、Python 3.8 |

The Flowering stage classification model is structured as a 7-layer neural network, shown in Fig. 4. The input includes the number of buds, blooming flowers, and withered flowers, as well as time information. The labels are manually recorded flowering stage. After passing through six hidden layers and the softmax function for classification, the final output is the predicted probability of each flowering stage class.

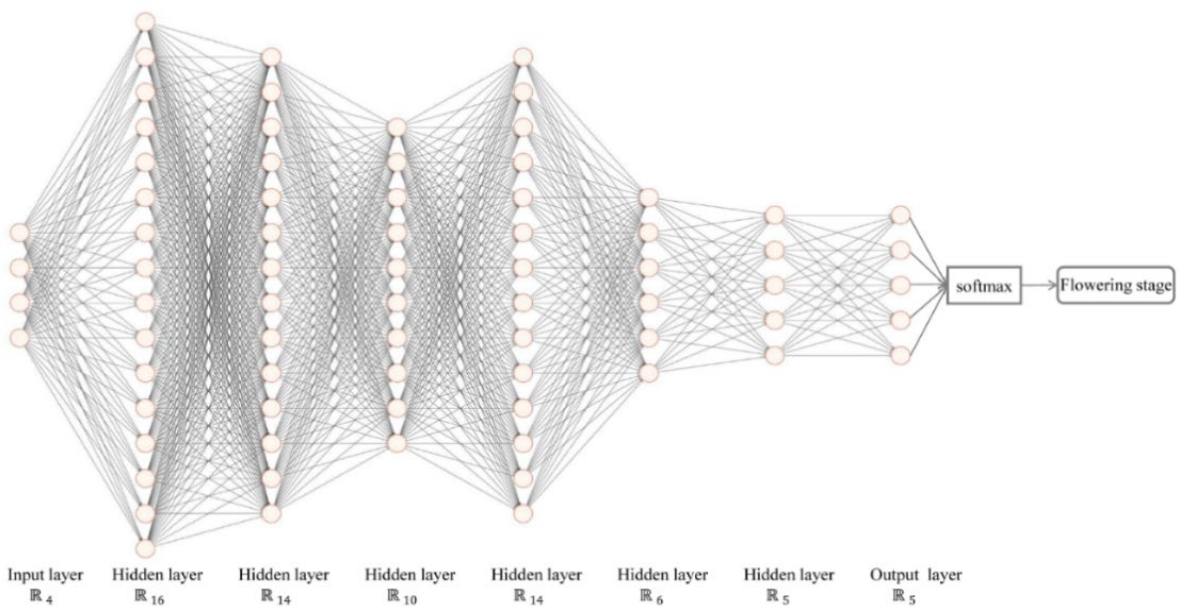

Fig. 4. The TFSC model (Tea flowering stage classification model).

The softmax function is calculated as:

$$\hat{y}_j = \frac{\exp(o_i)}{\sum_k \exp(o_k)} \quad (11)$$

Where $\hat{y}_j$ represents the predicted probability, $o_i$ is the unnormalized prediction for the $i_{th}$ output, and k is vector of predicted outputs. The softmax function ensures that the predicted outputs sum to 1, with each value in the range [0, 1].

The ReLU (Rectified Linear Unit) activation function is commonly used in artificial neural networks to introduce non-linearity and avoid issues such as gradient explosion and vanishing gradients. The ReLU function is defined as:

$$f(x) = \max(0, x) \quad (12)$$

*2.5.3 Model evaluation*

The accuracy is validated on the test set using the accuracy score function. The accuracy is calculated as:

$$ACC = \frac{TP+TN}{TP+TN+FP+FN} \quad (13)$$

Where TP, TN, FP, and FN represent true positives, true negatives, false positives, and false negatives, respectively.

## 3. RESULTS

### 3.1 TflosYOLO model performance and comparison

*3.1.1 TflosYOLO model performance for tea flower detection*

The model performance was evaluated using test dataset, and the results are summarized in Table. 4. The TflosYOLO model can accurately detect and locate tea flowers. For the three categories, the mAP50 was 0.874, precision was 0.802, recall was 0.854, and the F1 score was 0.827. The mAP50 for flower buds, blooming flower, and withered flowers all exceeded 0.82, with bud achieving the highest detection accuracy. The precision, recall, and F1 scores for bud and blooming flower were all above 0.80. While the performance for withered flower was slightly lower, it still exceeded 0.76. These results demonstrate that the model exhibits high accuracy and generalization capability. The model detection performance on one image is provided in Fig. S6, showing that TflosYOLO can accurately detect and locate tea flowers, even when they are obstructed by branches and leaves or when partial occlusions occur between flowers and bud. Additionally, mAP50 of TflosYOLO model on validation dataset set was 0.808(Table S5).

Table. 4. Performance of the TflosYOLO model based on test dataset.

| Class | Precision | Recall | $F_1$-score | mAP50 | mAP50-95 | Params /M | Model_size /M | GFloPs |
|---|---|---|---|---|---|---|---|---|
| all class | 0.802 | 0.854 | 0.827 | 0.874 | 0.696 | 15.8 | 30.4 | 34.9 |
| bud | 0.835 | 0.885 | 0.859 | 0.913 | 0.737 | | | |
| B flower | 0.801 | 0.867 | 0.833 | 0.881 | 0.685 | | | |
| W flower | 0.769 | 0.810 | 0.789 | 0.827 | 0.666 | | | |

*3.1.2 Evaluating the robustness of TflosYOLO model*

To assess the robustness and generalization ability of the TflosYOLO model, 34 additional test datasets were used, covering 26 tea accessions and 5 flowering stage datasets: IFS, EFS, MFS, LFS, TFS, along with unpruned tea plants and both backlight and frontlight conditions. The test results as shown in Fig. 5 presents the precision, recall, and mAP50 values for the TflosYOLO model across 34 additional test datasets.

The model performed slightly less effectively for accessions with very few flowers, such as EC1 and FY6, with mAP50 values reaching 0.74 or higher. For the majority of accessions, the mAP50 exceeded 0.8, and for several accessions, it was above 0.9. To prevent lengthiness, detailed results regarding the performance of TflosYOLO model on 34 test set have been provided in Table S6, 7, 8. The model performed best during the PFS (including EFS, MFS, LFS), with LFS showing the most accurate predictions, while IFS and TFS had the lowest accuracy (Fig. 5, Table S6, S8). The model performed slightly better on pruned tea plants compared to unpruned ones, but accuracy, recall, and F1 scores for both pruned and unpruned datasets exceeded 0.8. The model's performance under frontlight conditions was noticeably lower than under backlight, but the mAP50 remained above 0.8 under both conditions. In summary, accuracy of TflosYOLO model across most accessions, flowering stages, pruned and unpruned tea plants, and varying light conditions remained above 0.8, indicating high robustness and generalization capability.

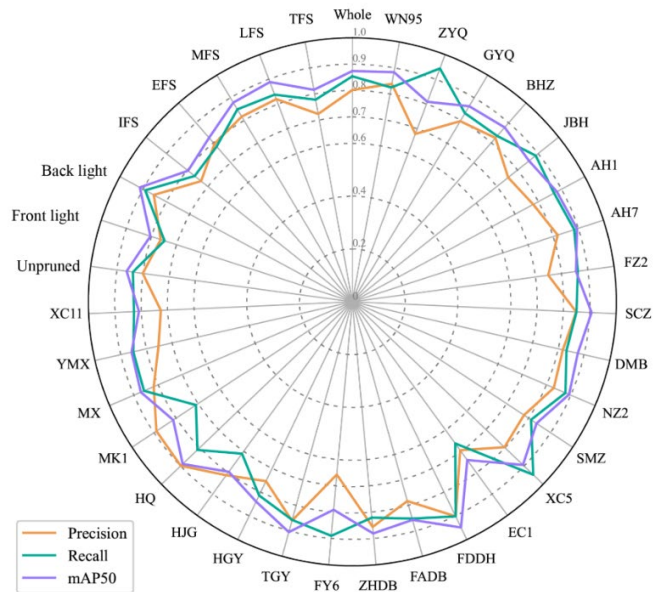

Fig. 5. The performance of TflosYOLO model on 34 additional test set.

### 3.1.3 Correlation Analysis

To further evaluate reliability of TflosYOLO model, correlation analysis was conducted using the R² coefficient. The correlation between the predicted flower count by TflosYOLO and the labeled flower count was computed based on the tea flower test dataset. The linear regression between the predicted flower count by TflosYOLO and the actual flower count (from labeled data) is shown in Fig. 6A. The correlation coefficient ($R^2$) for the predicted and actual flower count was 0.974, indicating a strong correlation between the predicted flower count and the actual count.

Additionally, the correlation between the predicted flower count and actual flower quantity levels from traditional manual surveys was analyzed. As shown in Fig. 6B, the predicted flower count and flower quantity level from traditional manual investigation across 26 accessions are basically consistent.

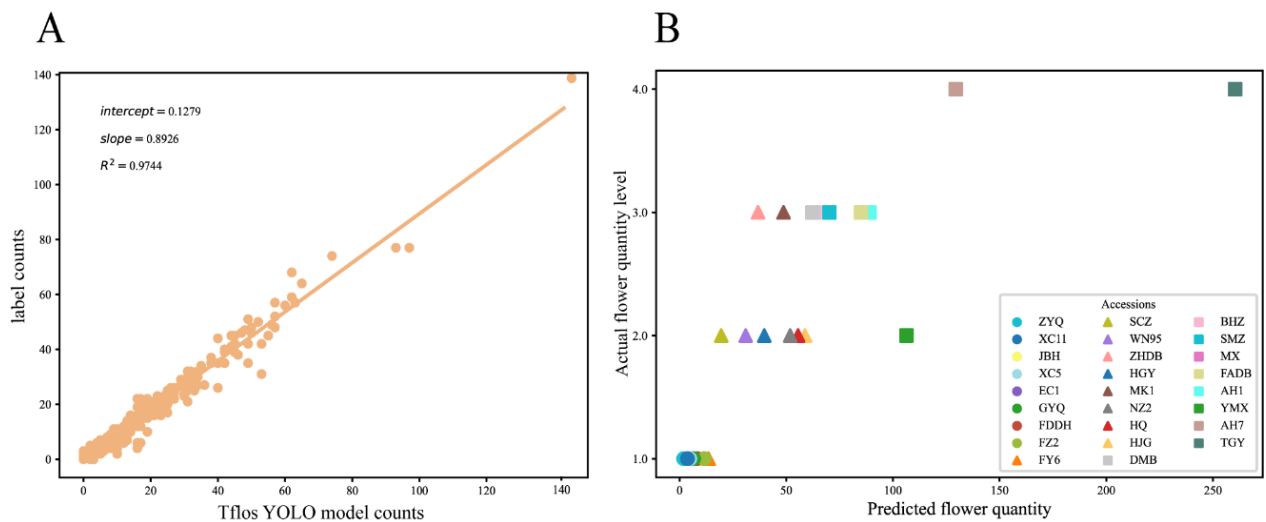

Fig. 6. The correlation between the predicted flower count by TflosYOLO and the actual flower count. (A) The linear regression between the predicted flower count and the actual flower count (from labeled data). (B) The flower quantity comparison between the predicted flower quantity and actual flower quantity levels from traditional manual surveys.

### 3.1.4 Ablation experiments of TflosYOLO model

This study used YOLOv5m as the baseline model and incorporated various improvements into TflosYOLO to improve model performance in different environmental conditions. The ablation experiment was conducted based on the validation dataset (Fig. 7). YOLOv5f modifies the depth and width of the YOLOv5 model, with depth and width multiplie of 0.33 and 0.75, respectively, placing it between YOLOv5s and YOLOv5m. Compared to the YOLOv5m model, YOLOv5f demonstrated increased accuracy with lower computational costs. Image enhancement (IE) led to significant improvements in precision, recall, F1-score, mAP50, and mAP50-95 compared to YOLOv5f. The addition of the Squeeze-and-Excitation (SE) module further increased the recall, F1-score, mAP50, and mAP50-95, with no change in the number of parameters, model size, or GFLOPs.

Table 5. The evaluation result of the ablation experiment.

| Model | Precision | Recall | F$_1$-score | mAP50 | mAP50-95 | Params /M | Model_size /M | GFloPs |
|---|---|---|---|---|---|---|---|---|
| YOLOv5m | 0.759 | 0.685 | 0.720 | 0.760 | 0.499 | 20.9 | 40.2 | 47.9 |
| YOLOv5f | 0.774 | 0.693 | 0.731 | 0.763 | 0.506 | 15.8 | 30.4 | 34.9 |
| YOLOv5f +IE | 0.795 | 0.712 | 0.751 | 0.793 | 0.490 | 15.8 | 30.4 | 34.9 |
| YOLOv5f+IE +SE | **0.792** | **0.727** | **0.760** | **0.808** | **0.523** | 15.8 | **30.4** | **34.9** |

Additionally, two test images were selected for the ablation experiment comparison: one under backlight with medium light intensity and the other under frontlight on a sunny day. The areas of interest are highlighted in white circles (Fig. 7). Under normal lighting conditions, the differences between models were minimal. However, with image enhancement, the TflosYOLO model correctly detect the flower calyx as a withered flower, whereas YOLOv5f misclassified it as a bud. Image enhancement and the SE attention module mitigated the issues caused by class imbalance, leading to more accurate detection of withered flowers. Under strong light and frontlight conditions, tea flower detection was interfered, with several objects missed by models in (A), (B) and (C) due to intense lighting. TflosYOLO showed superior performance under these conditions, detecting more buds and blooming flowers correctly.

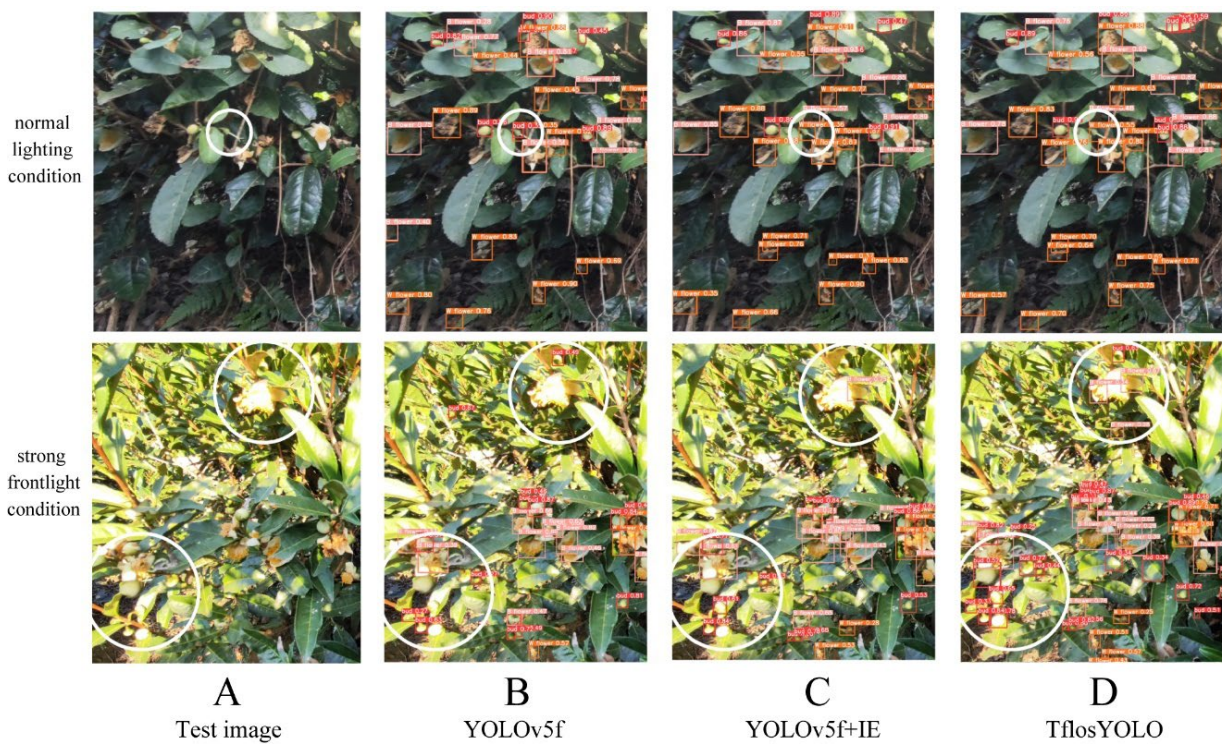

A — Test image  
B — YOLOv5f  
C — YOLOv5f+IE  
D — TflosYOLO

Fig. 7. Comparison of the detection effect of model improvement. (A) Test image. (B) YOLOv5f. (C) YOLOv5f+IE. (D) TflosYOLO, which include YOLOv5f+IE+SE.

In general, after the model improvements, the detection of withered flowers showed the greatest progress, fronted by blooming flowers, while improvements in bud detection were modest. TflosYOLO demonstrated noticeable improvements in detecting buds under strong light and also improved the detection of withered flowers. These model enhancements were beneficial in addressing challenges under strong light and frontlight conditions and were effective in mitigating class imbalance issues. Besides, the Squeeze-and-Excitation Networks contributed to model performance, and resistance to background noise.

*3.1.5 Comparative performance of YOLO algorithms for tea flower detection*

To compare the performance of the TflosYOLO model with other YOLO algorithms, we evaluated YOLOv5 (n/s/m/l/x), YOLOv7 (yolov7-tiny/yolov7/yolov7x), and YOLOv8 (n/s/m/l/x) models based on a validation dataset. The models were trained using the same parameters, and the results are summarized in Fig. 8A, B, Table S9. Compared to YOLOv5, YOLOv7, and YOLOv8, TflosYOLO performed better in detecting tea flowers, achieving higher precision, recall, and mAP50-95 while requiring fewer computational resources and having a model size between YOLOv5s and YOLOv5m. The table presents the average detection performance for the three classes-buds, blooming flower, and withered flowers.

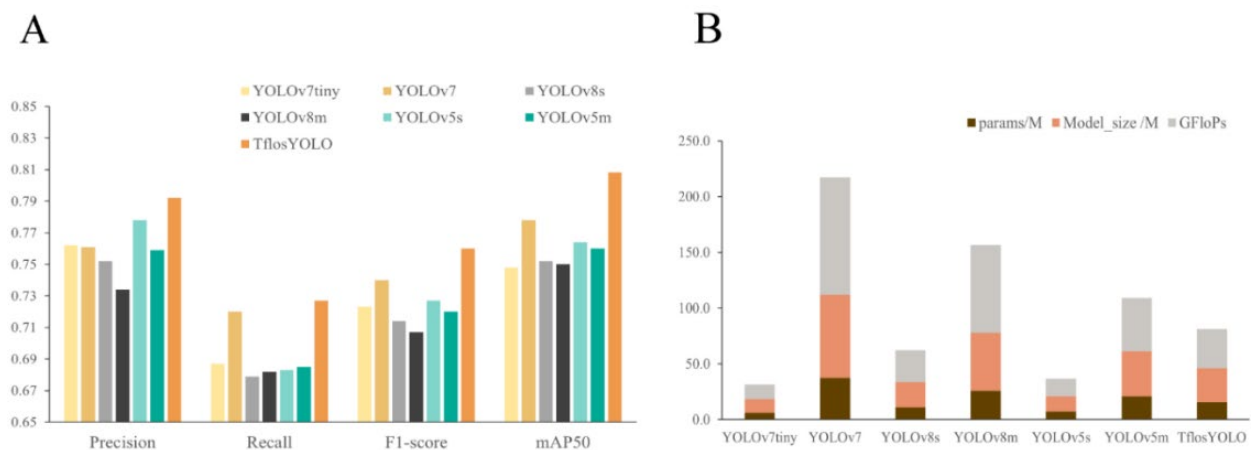

Fig. 8. Comparison of YOLOv5(n/s/m/l/x) & YOLOv7(tiny/yolov7/x) & YOLOv8(n/s/m/l/x) model performance.

(A) Comparison of model accuracy. (B) Comparison of model size and efficiency.

In conditions of bright light and front-light, TflosYOLO had a lower misidentification rate, accurately identifying flower buds and blooming flowers, while other models missed many flower buds or flowers under intense lighting (Fig. 9). In environments with moderate lighting, performances among models were similar (Fig. S7), but TflosYOLO correctly identified the flower calyx as a withered flower, whereas other models either failed to detect the calyx or misclassified it as a flower bud. In conclusion, the TflosYOLO model demonstrated superior performance in detecting tea flowers under both strong light and front-light conditions as it has higher accuracy, particularly for bud and withered flowers, while other models struggle with these conditions.

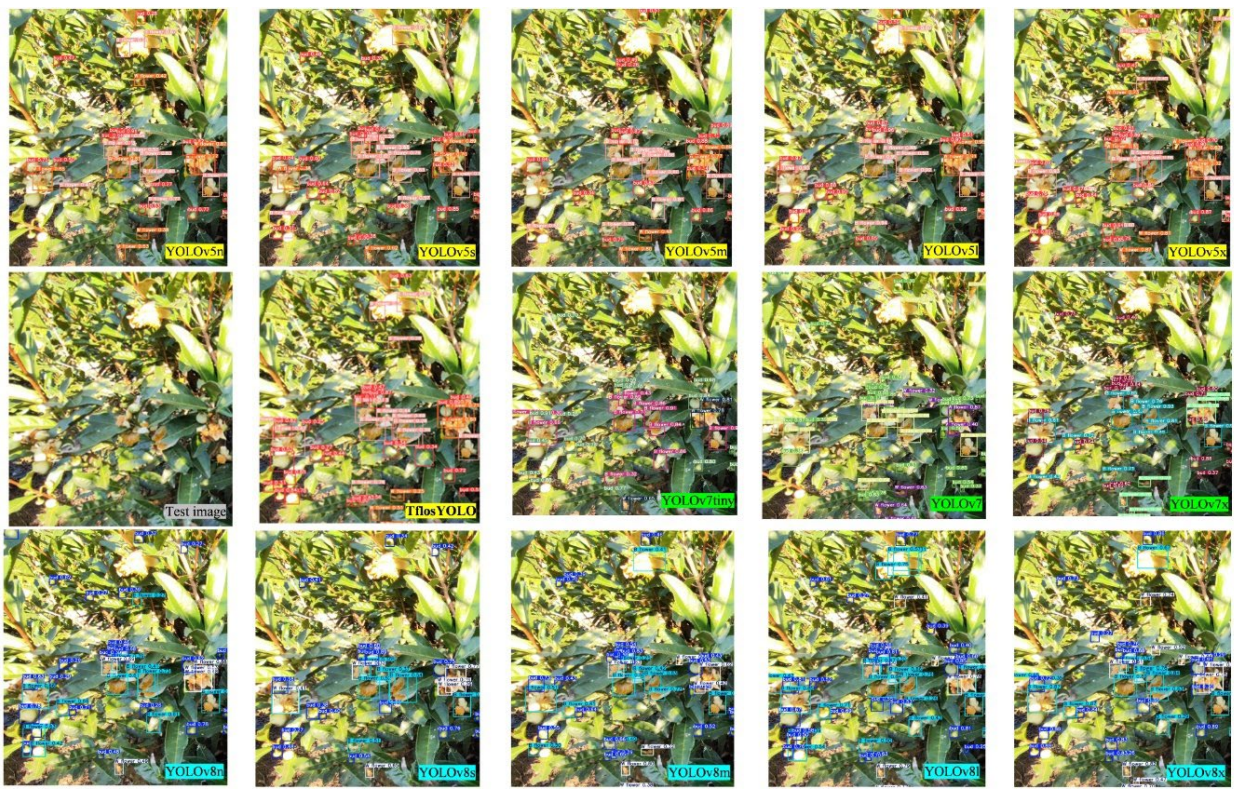

Fig. 9. Comparison of TflosYOLO with YOLOv5(n/s/m/l/x) & YOLOv7(tiny/yolov7/x) & YOLOv8(n/s/m/l/x) model under front-light condition on sunny day.

## 3.2 Evaluation of tea flowering stages classification model (TFSC)

The TFSC based on Artificial Neural Networks (ANN) achieved an accuracy of 0.738 and 0.899 on the validation dataset and test dataset respectively. The confusion matrix (Fig. 10) indicates that classification of the flowering stages is prone to misclassification between adjacent stages.

Specifically, there is frequent confusion between the EFS, MFS, and LFS, as the agricultural dataset contains a large number of intermediate periods and intermediate-type samples. Such misclassification is common in manual classification as well, especially between adjacent stages.

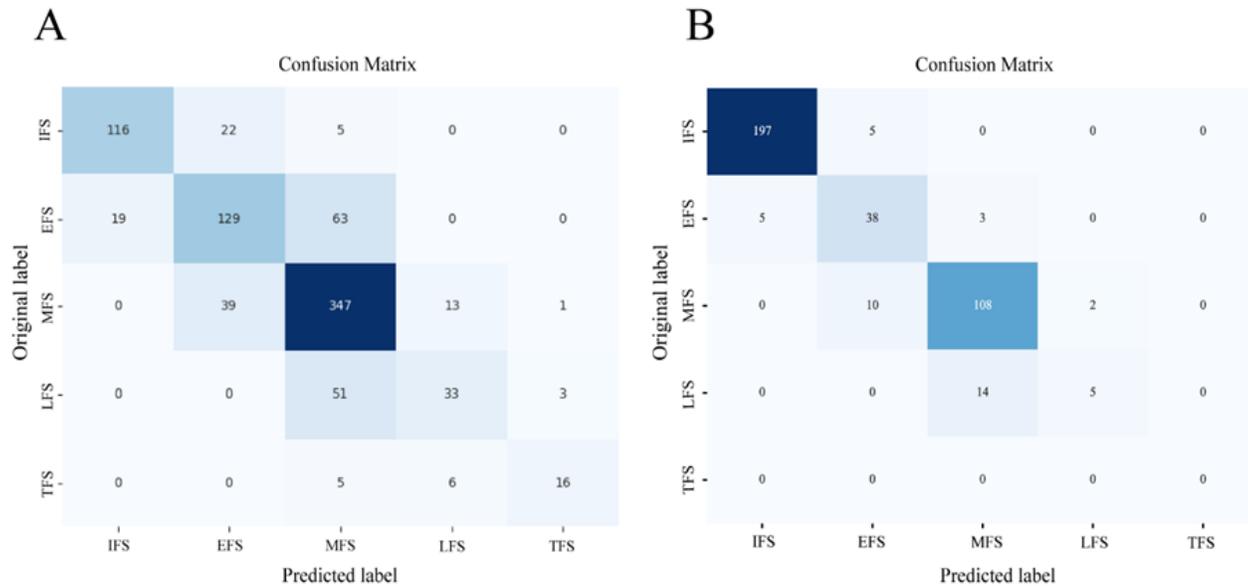

Fig. 10. The confusion matrix of predicted flowering stages and manual recorded flowering stages. (A) The confusion matrix based on validation dataset. (B) The confusion matrix based on test dataset.

## 3.3 Application of the TflosYOLO+TFSC Model in Flower Count and Flowering Period Estimation

The TflosYOLO+TFSC model was used to perform dynamic flower counting and flowering period estimation. We used time-series dataset constructed for observing tea flowering dynamics including 29 tea accessions and 5 flowering stages in 2023-2024, the composition of this dataset was summarized in Table S10, S11. The tea flowering observation dataset contains a total of 5,029 and 4345 images in 2023 and 2024.

*3.3.1 Monitoring of tea flowering dynamics with flowering stage information*

Using time-series images of 29 tea accessions in 2023, 2024 and the TflosYOLO + TFSC model model, we monitored the flowering dynamics and tracked the changes in flowering stages. The reference of flowering dynamics visualization was shown in Fig. 11. The tea flowering dynamics of other tea accessions in 2023 and 2024 are provided in Fig. S8, 9, 10. The flowering dynamics of different tea accessions exhibited distinct differences. In 2024, the flowering period of tea plants was generally later than in 2023. Moreover, based on the results, the relative early or late flowering of tea accessions is summarized in Table S12. With the exception of BHZ, the Flowering Stages predicted by the model aligned with those recorded manually.

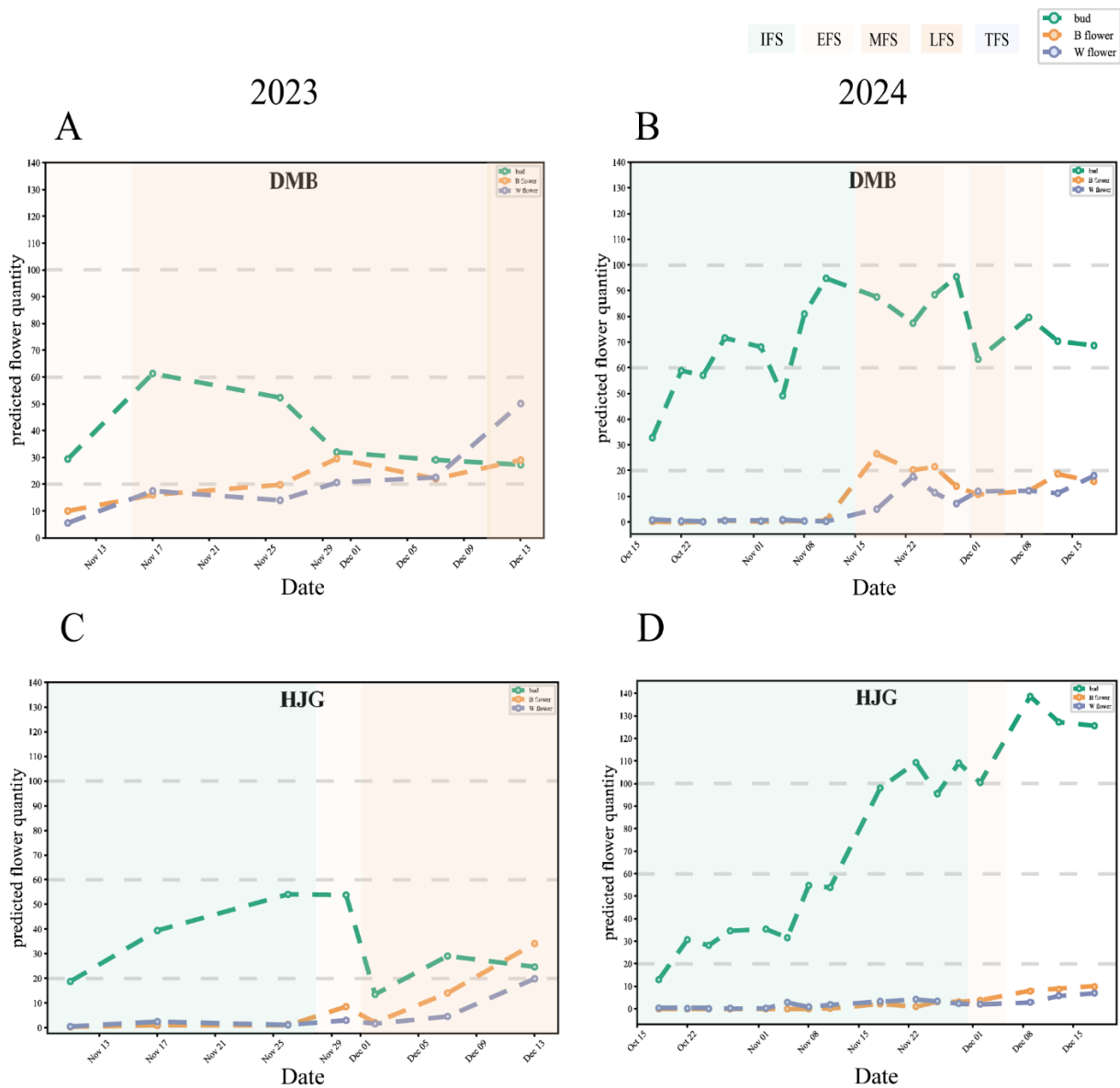

Fig. 11. Tea flowering dynamics and flowering period information for 2 accessions in November - December 2023 and October - December 2024. (A) DMB in 2023; (B) DMB in 2024; (C) HJG in 2023; (D) HJG in 2024.

*3.3.2 Estimation of flower quantity across different tea accessions, year and managements*

In this study, TflosYOLO was used to provide flower quantity data for each accession. The analysis and comparison of flower quantities across accessions were performed using data from the

2023-2024 Peak Flowering Stage (PFS) (Fig. 12A). Significant variability in flower quantity was observed across different tea accessions, and the flower quantity of the same accessions in 2023 and 2024 was relatively stable.

To further validate the robustness and reliability of the model, flower quantity under backlighting (BL) and frontlighting (FL) conditions was compared (Fig. 12B, C). The flower quantities under backlighting and frontlighting for same tea plants were similar, with no significant differences (p-value > 0.05). The results indicates that TflosYOLO model demonstrated stable performance under both lighting conditions, unaffected by lighting variations. Additionally, a significant difference in flower quantity was observed between pruned and unpruned tea plants. The flower quantity of both pruned and unpruned LJ43 tea plants was compared, and unpruned LJ43 plants exhibited significantly higher flower quantities than the pruned ones, with a p-value < 0.01 (Fig. 12D).

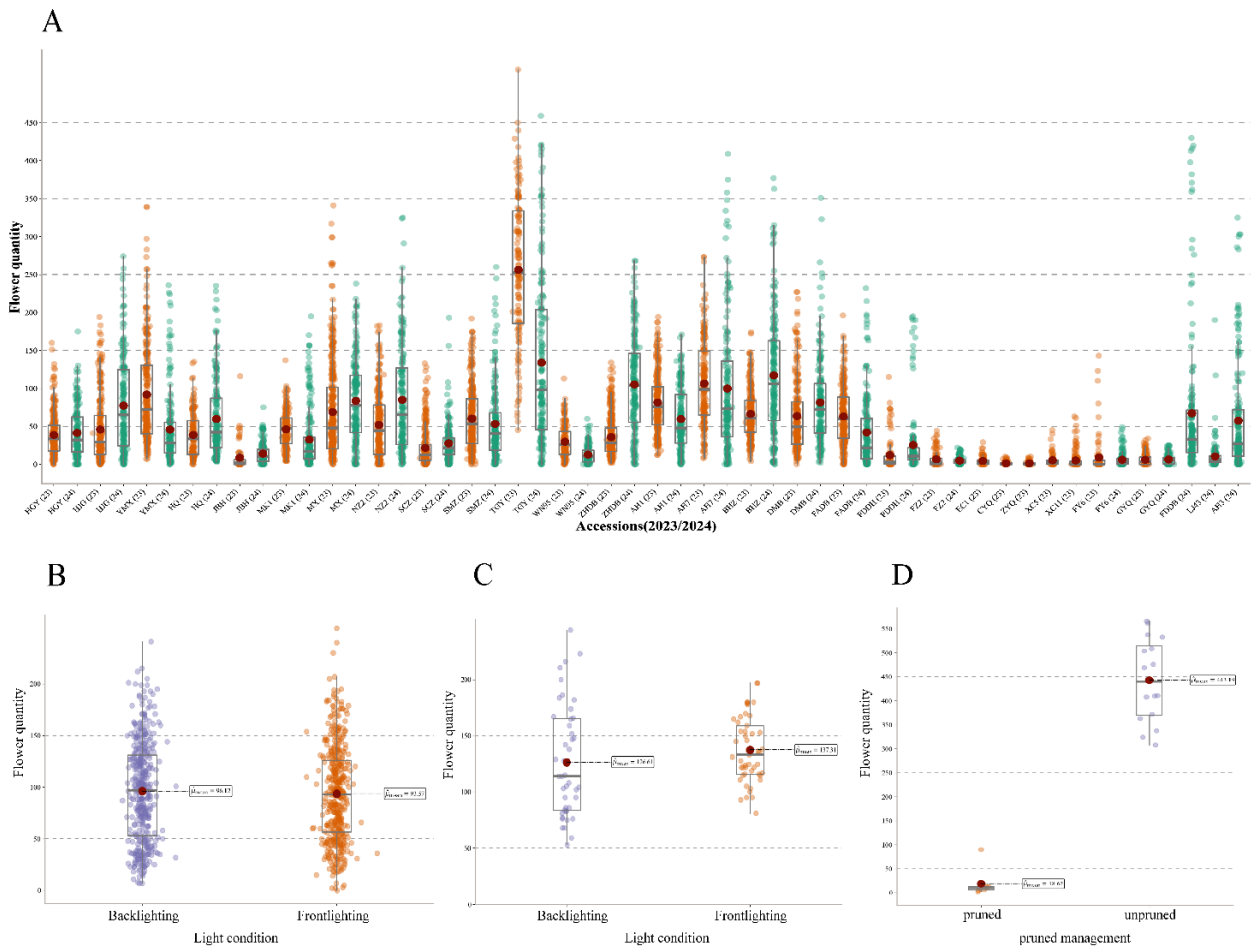

Fig. 12. Estimation of flower quantity across different tea accessions, year and managements. (A) Distribution of flower quantity across 29 accessions (2023, 2024). (B) Flower quantity under frontlighting and backlighting conditions for tea plants from the same plot; (C) Flower quantity of Jin Xuan tea plants under frontlighting and backlighting conditions. (D) Distribution of flower quantity of pruned / unpruned management LJ43.

*3.3.3 Distribution of flower quantity across different tea flowering stages*

Furthermore, TflosYOLO was used to analyzed the flower quantity for each flowering stage (IFS, EFS, MFS, LFS, TFS) separately, and flower quantity of 2 selected accessions were analyzed and shown in Fig. 13, data of accessions from other provinces is provided in Fig. S11.

The flower quantity during different flowering stages vary significantly. While most tea accessions do not show significant differences in flower quantity between 3 PFS (EFS, MFS, LFS), significant differences in flower quantity were observed among IFS, PFS, and TFS.

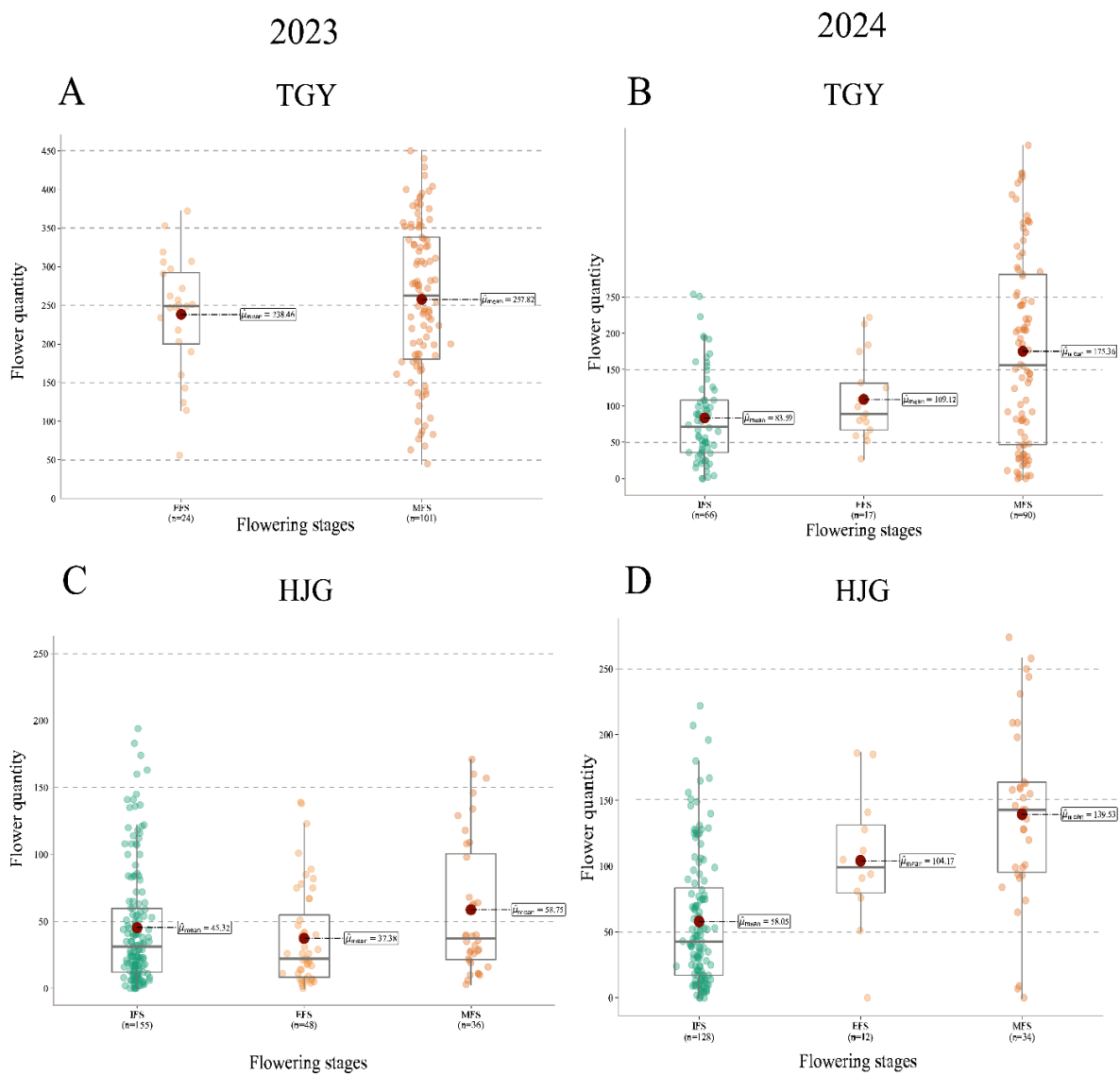

Fig. 13. Flower quantity data for different flowering stage (IFS, EFS, MFS, LFS, TFS) across 2 accessions in 2023 and 2024. (A) TGY 2023; (B) TGY 2024; (C) HJG 2023; HJG 2024.

## 4. DISCUSSION

**Importance of Datasets:** Agricultural datasets typically present challenges such as significant background noise and small object sizes, making the model performance very different from the evaluations done using datasets like COCO. For example, in this study, YOLOv5s outperformed the more computationally intensive YOLOv5l x and even YOLOv8. In the training and construction of deep learning models, such as YOLO, the representativeness and diversity of the dataset may be more crucial than improvements in the model architecture. The performance of model can vary significantly across different accessions. Therefore, achieving good results on a single dataset does not guarantee consistent performance across all scenarios, and it is essential to test the model in different environments and with different accessions. Moreover, we have validated the feasibility of employing the YOLOv5 computer vision model in complex field environments, demonstrating its applicability across different tea varieties. This validation allows us to assess the extent to which varietal differences influence model performance.

In this study, incorporating attention mechanisms such as SE, CBAM, and CEA led to significant improvements in cases with insufficient datasets, while their impact was less pronounced when the dataset was sufficiently large. Moreover, the composition of the dataset clearly affects the model performance. For instance, the predictions for the PFS (including EFS, MFS, LFS) were the most accurate, particularly for the LFS, while performance during IFS and TFS was poorer. This is likely due to the training dataset predominantly consisting of images from the PFS.

**Model construction for Basic Data**: For relatively simple datasets, such as the flowering stage data in this study, a simple artificial neural network suffices for classification tasks. After designing and comparing different network architectures in this study, it was found that increasing the complexity of the model does not lead to improvements in performance.

**Consideration of agronomic characteristics in quantifying different crop Traits:** When quantifying agronomic traits in crops, it is essential to account for specific agronomic characteristics. For example, tea flower quantity is greatly influenced by light exposure, and there are substantial variations in flower quantity across different tea plant of the same row. Thus, it is important to collect a sufficient number of images from various locations within the field. Additionally, tea accessions exhibit differences in morphology-ranging from small trees to shrubs and the significant image disparities between pruned and naturally grown trees require models with high generalization and robustness.

**Influence of plant size and weather on Tea Flower Quantity:** Flower quantity is strongly correlated with the size of the tea plant. To compare flower quantities across different accessions, it is important to ensure that the comparisons are made between plants of similar size and management practices. Additionally, tea flower quantity is influenced by weather conditions. Due to climatic differences between 2023 and 2024, the flowering dynamics of the same accession varied significantly and flowering period was generally later in 2024 than in 2023, as the extreme low temperatures in November and December 2023 were lower than those in November and December 2024. In the future, it would be valuable to combine tea flowering data with meteorological data to analyze the dynamics of tea plant flowering. Additionally, the observed flower quantity is significantly affected by both flowering period and the timing of image acquisition. Consequently, observations made over a short time frame may not accurately reflect the true flowering dynamics.

**Comparation with previous tea flower studies:** Although previous tea flower studies constructed by manual survey involved fewer accessions, the overall flower quantity and flowering stage align with our findings. For instance, the flower quantity of accessions like MX and TGY was consistently high across different studies, and HJG displayed relatively high quantity.

## 5. CONCLUSIONS

This study proposes an effective framework for quantifying tea flowering, comprising the TflosYOLO model and TFSC model. Compared to traditional manual surveys and observations, this framework is more efficient and accurate. The TflosYOLO model demonstrates the ability to accurately detect tea flowers under various conditions, including different tea accessions, flowering stage, pruning practices, and lighting conditions. Its high robustness and generalization capability render it the only model currently suitable for detecting and counting tea flowers, achieving state-of-the-art (SOTA) performance in this domain. Additionally, TFSC model consistently demonstrates an accuracy exceeding 0.73 across different years, indicating its high generalizability. TflosYOLO combined with TFSC model enable accurate estimation of flower count and flowering period across different accessions.

Based on TflosYOLO combined with TFSC model, we found that there are differences in the flowering dynamics of various tea accessions. Accessions that are genetically related tend to exhibit more similar flower quantities and blooming periods. The flowering quantity and flowering period of the same accession can vary between different years due to changes in climate and management practices.

## ACKNOWLEDGMENTS

**Author contributions:** QM contributed to conceptualization, data collection, methodology, data analysis, writing- review & editing. JC contributed to conceptualization, writing- review & editing. MY contributed to review, Supervision. PY contributed to data collection. CM contributed to conceptualization.

**Funding:** This work was supported by the National Key Research and Development Program of China (2021YFD1200203), the Major Project of Agricultural Science and Technology in Breeding of Tea Plant Variety in Zhejiang Province (2021C02067-1), the Guangxi Key Research and Development Program (AB23026086), and the Fundamental Research Fund for Tea Research Institute of the Chinese Academy of Agricultural Sciences (1610212023003).

**Competing interests:** The authors declare that they have no competing interests.

**Data Availability:** Part code and some images of tea flowers used in this study are available at the GitHub repository: https://github.com/sufie-mi/tea-flower-model.